\documentclass{article}

\usepackage[preprint,nonatbib]{neurips_2023}

\usepackage[utf8]{inputenc} 
\usepackage[T1]{fontenc}    
\usepackage{hyperref}       
\usepackage{url}            
\usepackage{booktabs}       
\usepackage{amsfonts}       
\usepackage{nicefrac}       
\usepackage{microtype}      
\usepackage{graphicx}
\usepackage{subfigure}
\usepackage{wrapfig}

\usepackage[table,xcdraw,dvipsnames]{xcolor} 
\usepackage{color}
\definecolor{citecolor}{RGB}{66,168,235}
\definecolor{linkcolor}{RGB}{255,0,0}

\usepackage{hyperref}
\hypersetup{colorlinks=true,citecolor=citecolor,linkcolor=linkcolor,urlcolor=Thistle}

\title{Scalable Mask Annotation for Video Text Spotting}

\author{
Haibin He$^{1}$, \quad
Jing Zhang$^{2}$, \quad
Mengyang Xu$^{1}$, \quad
Juhua Liu$^{1}$, \quad
Bo Du$^{1}$, \quad
Dacheng Tao$^{2}$ \quad
\vspace{1 mm}
\\
\vspace{1 mm}
\textsuperscript{1}Wuhan University, China \\
\textsuperscript{2}The University of Sydney, Australia \\
\tt\small \{haibinhe, xumengyang, liujuhua, dubo\}@whu.edu.cn\\ 
\tt\small jing.zhang1@sydney.edu.au, dacheng.tao@gmail.com
}

\begin{document}

\maketitle

\begin{abstract}
  Video text spotting refers to localizing, recognizing, and tracking textual elements such as captions, logos, license plates, signs, and other forms of text within consecutive video frames. However, current datasets available for this task rely on quadrilateral ground truth annotations, which may result in including excessive background content and inaccurate text boundaries. Furthermore, methods trained on these datasets often produce prediction results in the form of quadrilateral boxes, which limits their ability to handle complex scenarios such as dense or curved text. To address these issues, we propose a scalable mask annotation pipeline called \textbf{SAMText} for video text spotting. SAMText leverages the SAM model \cite{Related31} to generate mask annotations for scene text images or video frames at scale. Using SAMText, we have created a large-scale dataset, SAMText-9M, that contains over 2,400 video clips sourced from existing datasets and over 9 million mask annotations. We have also conducted a thorough statistical analysis of the generated masks and their quality, identifying several research topics that could be further explored based on this dataset. The code and dataset will be released at \href{https://github.com/ViTAE-Transformer/SAMText}{SAMText}.
\end{abstract}

\section{Introduction}
Text spotting has garnered increased attention from both academia and industry due to its wide use in vision-based applications, such as visual translation \cite{zhang2020empowering}, autonomous driving \cite{Related03}, and video retrieval \cite{Related01}. With the availability of finely annotated public datasets \cite{Related05, Related06, Related07} and the rapid development of deep learning techniques, remarkable progress has been made in text spotting for static images. Typically, text spotting contains two sub-tasks~\cite{long2021scene}, i.e., text detection~\cite{liao2020real,du2022i3cl,ye2022dptext} and text recognition~\cite{zhan2019esir,he2022visual}. Prior CNN-based methods \cite{Related08, Related09, Related10} have modeled these two sub-tasks in an end-to-end framework that first detects text regions and then recognizes text within those regions. Other methods \cite{Related12, Related13} have simplified the pipeline by processing detection and recognition simultaneously with a single DERT framework \cite{Related11} and have achieved satisfactory results in spotting arbitrary-shaped text in images. In comparison to text spotting in static images, video text spotting is even more challenging as it involves three sub-tasks, namely detection, recognition, and tracking. While TransVTSpotter \cite{Related14} is the first to introduce the Transformer \cite{Related15} in video text spotting, CoText \cite{Related16} has developed a trainable end-to-end framework with three network heads to address all three sub-tasks simultaneously. Furthermore, TransDERT \cite{Related17} has proposed a simple pipeline without the need for multiple models and hand-crafted strategies such as non-maximum suppression and track post-processing, which achieves real-time video text spotting.

Large-scale and high-quality datasets are essential to fuel the development of deep learning algorithms. ImageNet \cite{Related18}, for instance, has contributed significantly to the training of popular image models such as ResNet \cite{Related20} and ViT \cite{Related21}, leading to remarkable results in tasks like classification, object detection, and segmentation. Similarly, LAION-5B \cite{Related22} has provided billions of finely annotated image-text pairs, which have made great contributions to image-text cross-domain tasks. In video text spotting, several public datasets have been established to drive the progress of text identification in videos. For instance, ICDAR2015 (Text in Videos) \cite{Related26} contains 49 video clips, where the location of words in the video is labeled using oriented bounding boxes (OBB). RoadText-1K \cite{Related27} collects 1,000 driving videos for driver assistance and self-driving systems, with the position of text annotated using horizontal bounding boxes (HBB). LSVTD \cite{Related28} includes 129 videos with more than three kinds of text languages, annotated using polygon coordinates that represent text location. BOVText \cite{Related14} introduces a large-scale, bilingual, open-world video text dataset, providing 2000+ videos with various scenarios, such as Life Vlog, Driving, and Movie. Finally, DSText \cite{Related29} establishes a dense and small text video dataset, including 100 videos, where both BOVText and DSText annotate videos with OBB.

Annotating vast amounts of data, particularly videos, requires substantial labor costs. Consequently, current video text spotting datasets utilize quadrilateral bounding boxes for text localization instead of more precise labels such as segmentation masks. However, training models with such coarse annotations significantly limits their performance and applicability. Fine text annotations, such as segmentation masks, can bring several benefits. First, the use of fine annotations can result in significant improvements in detection and recognition performance, as demonstrated in object detection tasks \cite{Related30} and scene image text spotting tasks \cite{Related10}. Second, models trained on datasets with fine annotations can be applied to a wider range of scenarios, such as curved text spotting. Third, fine-annotated datasets can be utilized in other related tasks such as video text segmentation, and curved text editing and removal. Therefore, it is critical to create a large-scale video dataset with fine annotations at a low cost to advance the development of video text spotting and beyond.

\begin{figure}[t]
  \centering
    \includegraphics[width=1\linewidth]{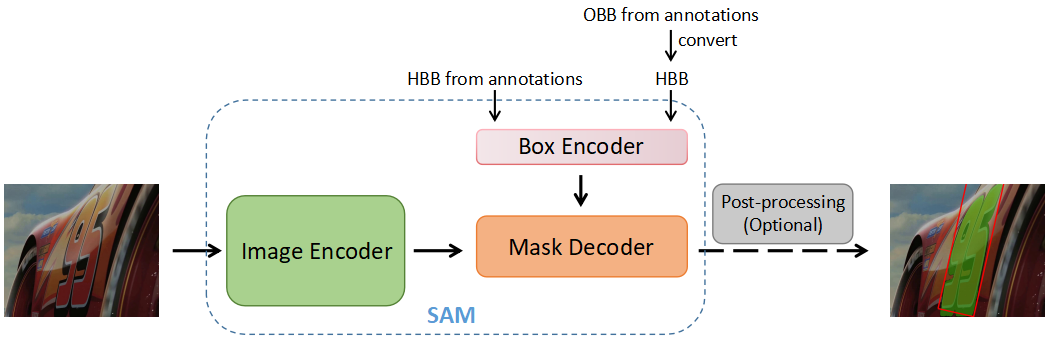}
    \caption{Overview of the SAMText pipeline that builds upon the SAM approach \cite{Related31} to generate mask annotations for scene text images or video frames at scale. The input bounding box may be sourced from existing annotations or derived from a scene text detection model.}
\label{fig:1}
\end{figure}

Recently, Meta AI Research has introduced a novel foundation model for image segmentation, known as the Segment Anything Model (SAM) \cite{Related31}. This model is trained on over one billion masks from 11 million images and can effectively segment objects based on point or box prompts. In this report, we propose SAMText, a scalable mask annotation pipeline for video text spotting that takes advantage of SAM's remarkable segmentation capabilities to generate mask annotations for scene text images or video frames at scale. Specifically, we collect five datasets, namely ICDAR15 \cite{Related26}, RoadText-1K \cite{Related27}, LSVTD \cite{Related28}, BOVText \cite{Related14}, and DSText \cite{Related29}, then leverages the SAM model \cite{Related31} to generate segmentation annotations. Using SAMText, we have created a large-scale dataset, SAMText-9M, that contains over 2,400 video clips sourced from existing datasets and over 9 million mask annotations. Moreover, we have conducted a thorough statistical analysis of the generated masks and their quality, identifying several research topics that could be further explored based on this dataset.

\section{Dataset}
\subsection{The SAMText Pipeline}
As illustrated in Figure~\ref{fig:1}, given an input scene text image or video frame, we begin by extracting the bounding box coordinates from existing annotations (or derived from a scene text detection model). If the boxes are oriented, we calculate their minimum bounding rectangle to obtain the HBB, which is then used as the input prompt for SAM \cite{Related31} to obtain mask labels. Upon obtaining the mask for each text instance, it may be necessary to perform post-processing to ensure its connectivity. In particular, if a mask comprises several segments, it may be desirable to derive a minimum enclosing mask as an optional step in order to achieve a more cohesive representation. Furthermore, optical flow estimation~\cite{teed2020raft,xu2022gmflow} can also be utilized to enhance the accuracy of the generated masks and ensure temporal consistency.

We select five widely used video text spotting datasets, namely ICDAR15 \cite{Related26}, RoadText-1K \cite{Related27}, LSVTD \cite{Related28}, BOVText \cite{Related14}, and DSText \cite{Related29}, that comprise either HBB annotations or OBB annotations for text instances across consecutive frames. Then, we employ the SAMText pipeline described above to generate segmentation annotations for each HBB or OBB instance in the training and validation datasets. In summary, we create a large-scale dataset, SAMText-9M, that consists of over 2,400 video clips and over 9 million mask annotations. Furthermore, to unify the annotations across the different datasets, we categorize the transcription annotations into three major groups: Alphanumeric (e.g., English, French, and Spanish), Non-alphanumeric (e.g., Chinese), and Others (i.e., without transcription annotation). Table~\ref{table_1} presents an overview of the statistical information for the five source datasets as well as for the SAMText-9M dataset. 

\begin{table}[t]
\setlength{\tabcolsep}{0.15cm}
\caption{Comparison of SAMText-9M with existing video text spotting datasets.}
\centering
\begin{tabular}{cccccc}
\hline
Dataset & \#Video & \#Frame & \#Text & Annotation Type & Resolution\\
\hline
ICDAR15 \cite{Related26}  &25 &13K &71K &OBB  & 480$\times$720 $\sim$ 960$\times$1,280\\
RoadText-1K \cite{Related27}  &700 &210K &937K &HBB  & 720$\times$1,280\\
LSVTD \cite{Related28}  &89 &62K &603K &OBB  & 640$\times$368 $\sim$ 1,440$\times$2560\\
BOVText \cite{Related14}  &1,540 &1.3M &6.7M &OBB  & 240$\times$432 $\sim$ 2,160$\times$5,760\\
DSText \cite{Related29}  &50 &12K &945K &OBB  & 720$\times$1,280 $\sim$ 2,160$\times$3,840\\
\hline
\textbf{SAMText-9M} &2,404 &1.6M &9.2M &Mask + OBB & 240$\times$432 $\sim$ 2,160$\times$5,760\\
\hline
\end{tabular}
\label{table_1}
\end{table}

\subsection{Dataset Statistics}
After obtaining the mask annotations for all five datasets, we perform a statistical analysis of the generated masks and their quality. It comprises four key aspects: 1) the size distribution of the generated masks for each dataset, 2) the distribution of Intersection over Union (IoU) values between the generated masks and the ground truth bounding boxes for each dataset, 3) the distribution of coefficient of variation (CoV) of mask size of the tracked instance across consecutive frames for all the datasets, and 4) the spatial distribution of the generated masks for each dataset.

\textbf{The size distribution.} We conduct an analysis of the size distribution of the masks generated for each video dataset. Specifically, we count the mask sizes for all instances within each dataset. Notably, we observe a significant deviation in the size distribution of the masks generated from the BOVText dataset as compared to the other four datasets. Therefore, we perform separate visualization for BOVText. For ICDAR15, RoadText-1K, LSVTD, and DSText, we calculate the number of masks with sizes less than 10,000 pixels with an interval of 400, whereas, for BOVText, we consider masks with sizes less than 80,000 pixels with an interval of 400.

\textbf{The IoU distribution.} In order to investigate the ratio of foreground to background in each instance, we calculate the IoU between each generated mask and its corresponding ground truth bounding box, and establish the IoU distribution for each dataset with a fixed interval of 0.1.

\textbf{The CoV distribution.} To investigate the variations in mask sizes of individual instances across consecutive frames, we conduct an analysis of the distribution of CoV (i.e., the ratio of standard deviation to mean) of mask sizes for tracked instances. To accomplish this, we utilize the tracking identity of each instance and collected all masks belonging to the same instance from consecutive frames of each video clip. We then calculate the standard deviation and mean of their mask sizes, repeating this process for each sequence of tracked masks across all video clips in the five datasets. Finally, we establish the CoV distribution with a fixed interval of 0.1.

\textbf{The spatial distribution.} To gain insight into the spatial distribution of the generated masks, we conduct an analysis of their spatial locations for each dataset. Given the varying resolutions of the videos in the dataset, we resize the masks to a standardized resolution of 1,280 $\times$ 1,280 and superimpose the binary masks together. We then normalize the resulting accumulated mask maps and represent them as pseudo-color heatmaps for visual interpretation.

\begin{figure}[t]
\centering
\setlength{\tabcolsep}{0.05cm}
\begin{tabular}{ccccc}
  \includegraphics[width=2.6cm,height=2.6cm]{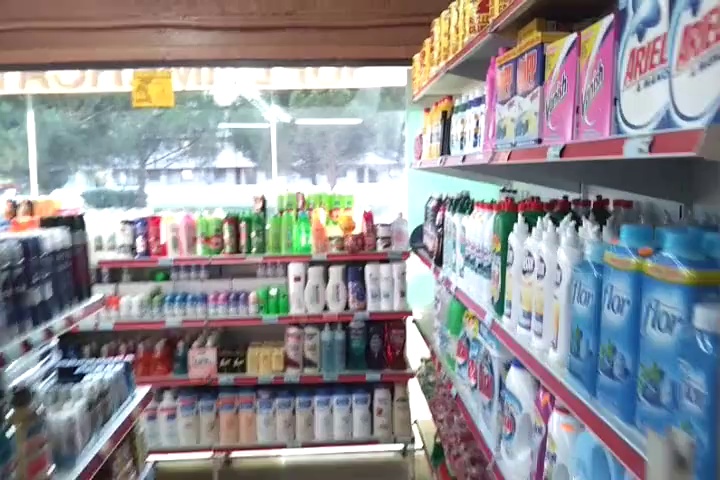} 
  &\includegraphics[width=2.6cm,height=2.6cm]{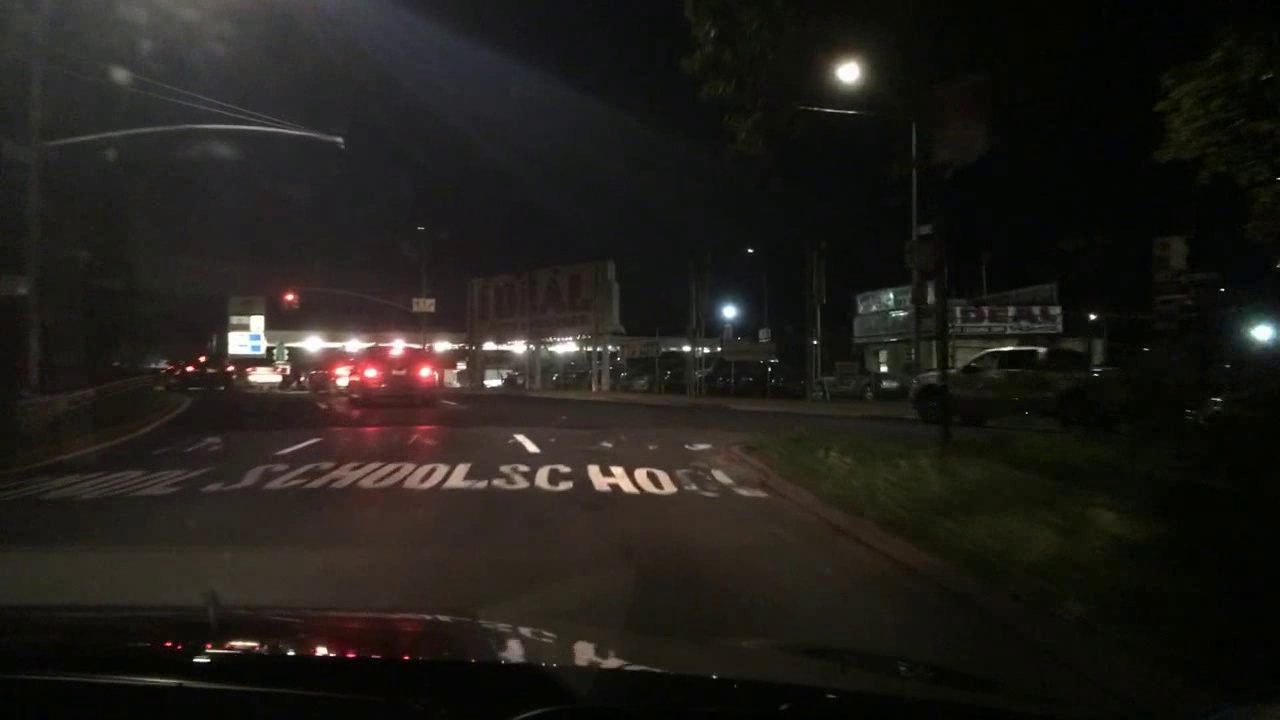}
  &\includegraphics[width=2.6cm,height=2.6cm]{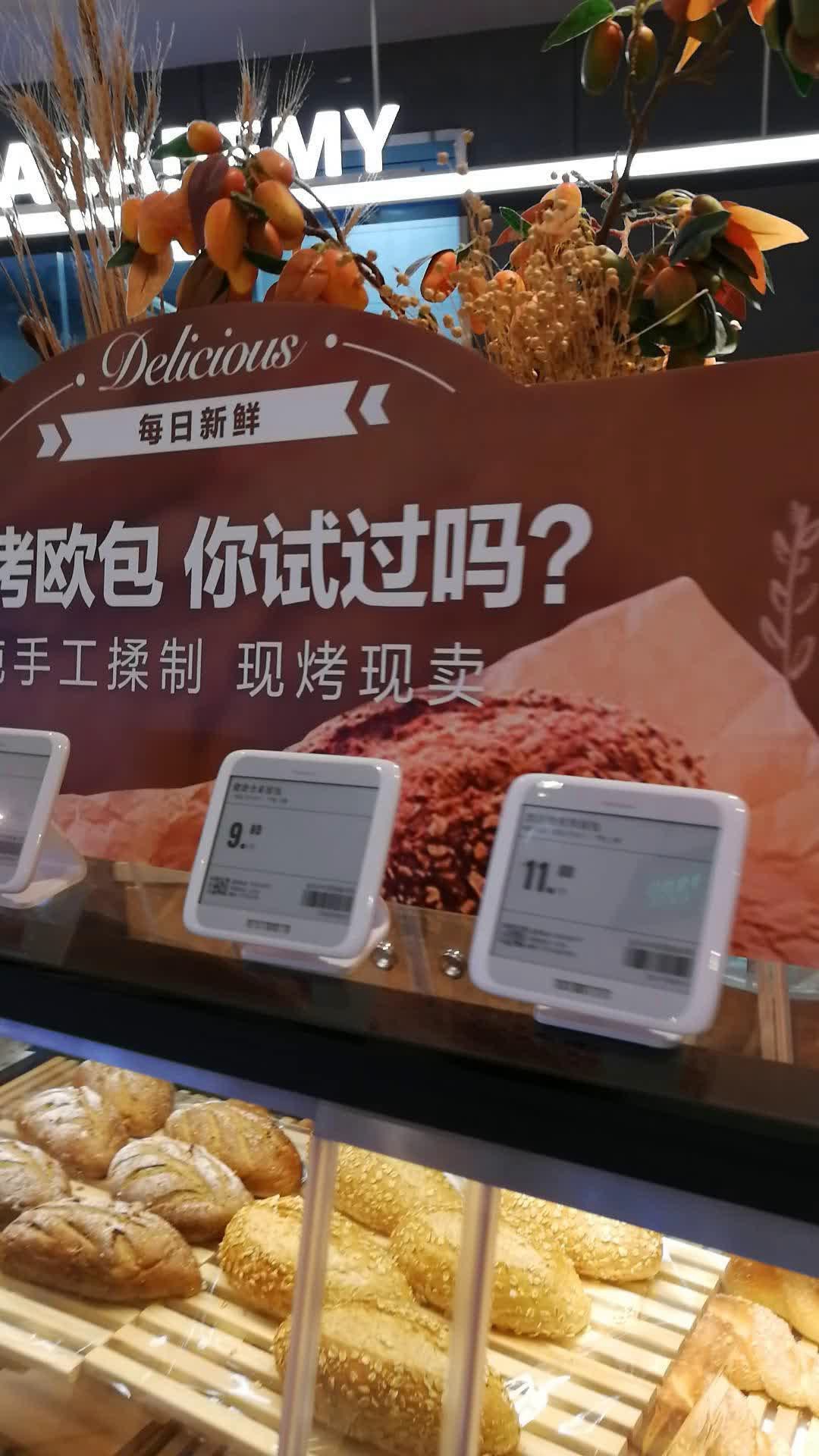}
  &\includegraphics[width=2.6cm,height=2.6cm]{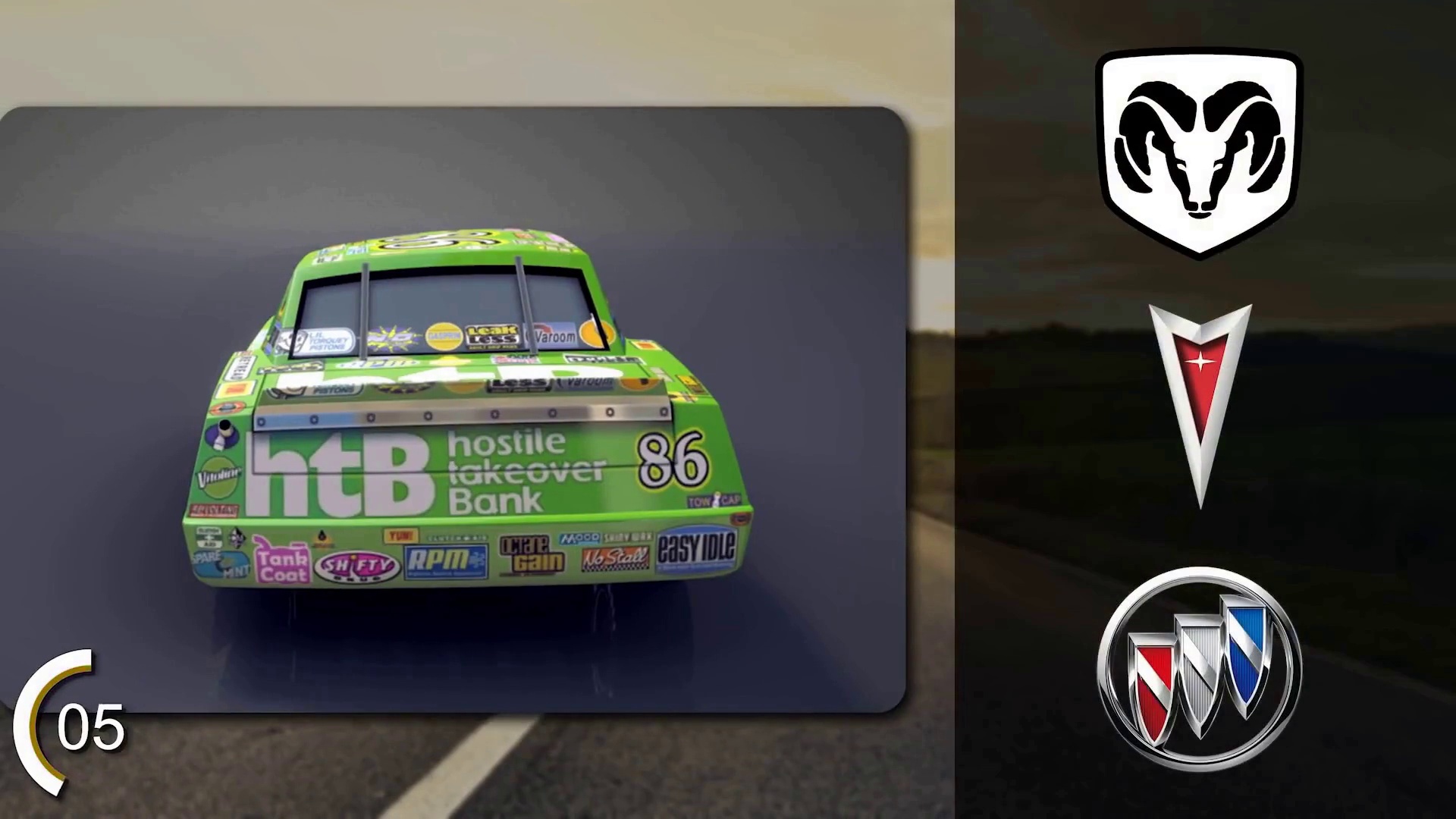}
  &\includegraphics[width=2.6cm,height=2.6cm]{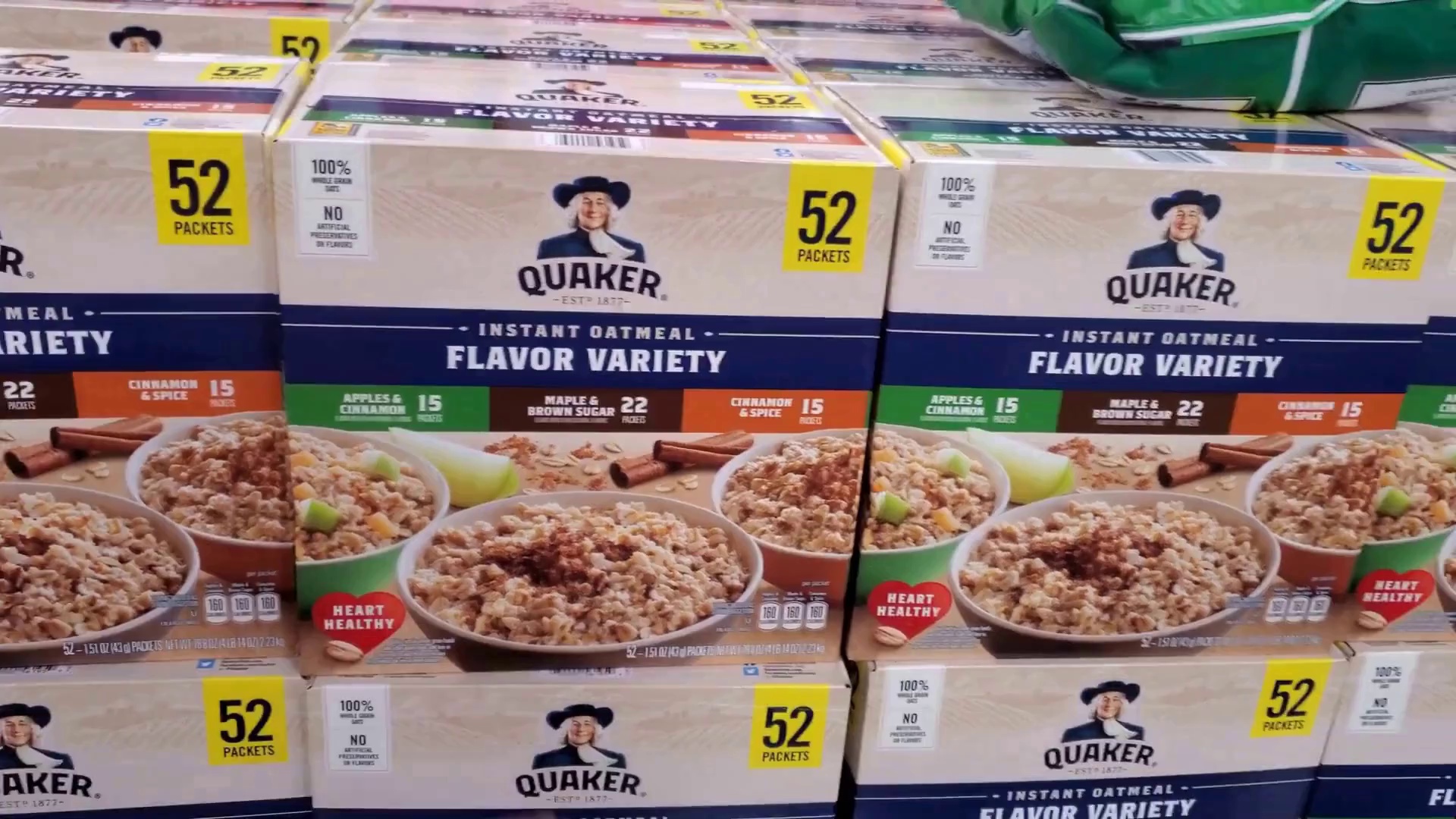} 
  \\ 
  \includegraphics[width=2.6cm,height=2.6cm]{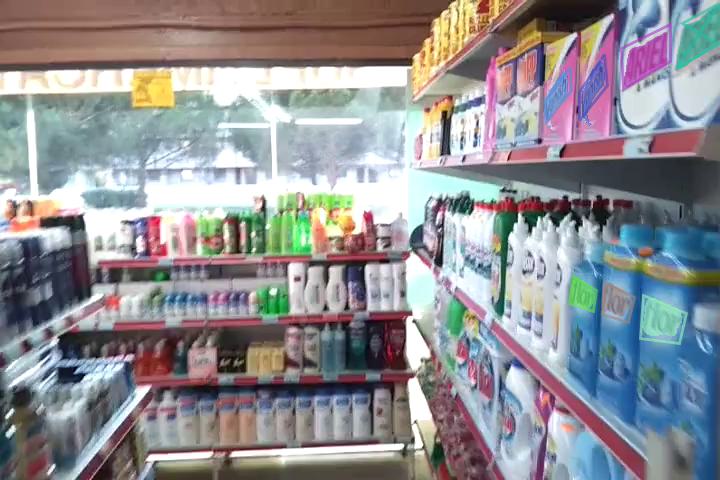}
  &\includegraphics[width=2.6cm,height=2.6cm]{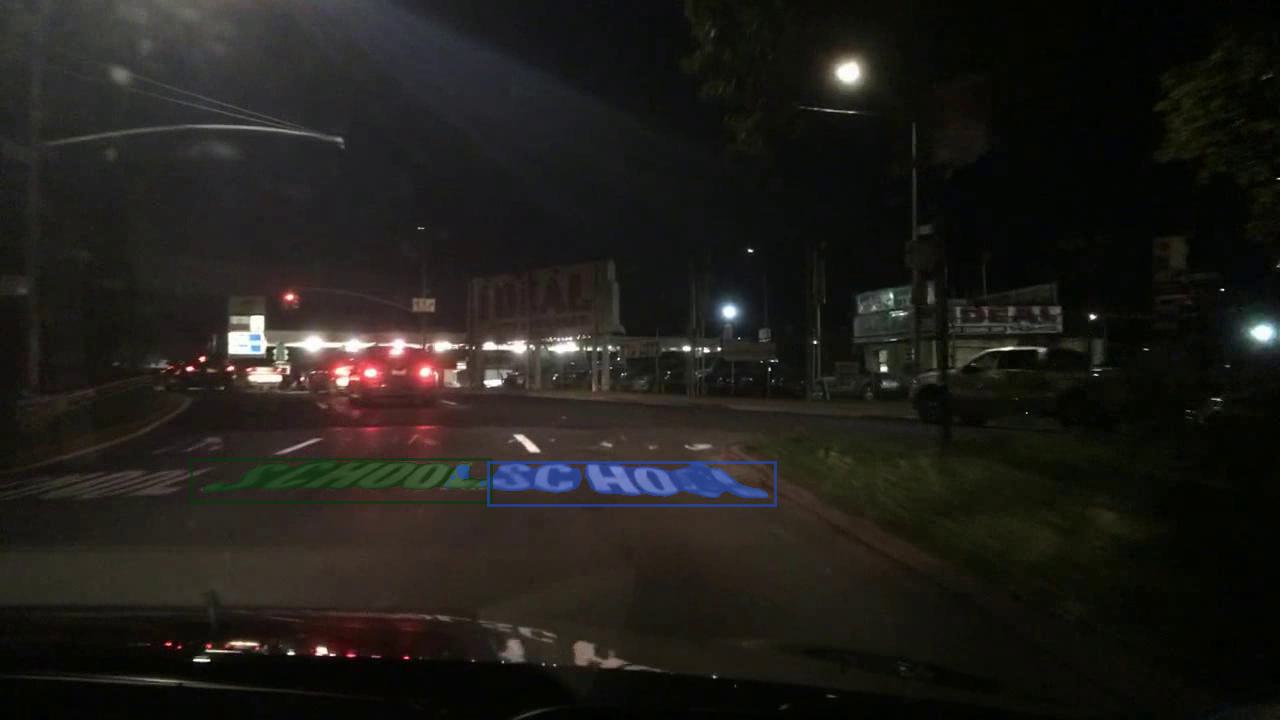}
  &\includegraphics[width=2.6cm,height=2.6cm]{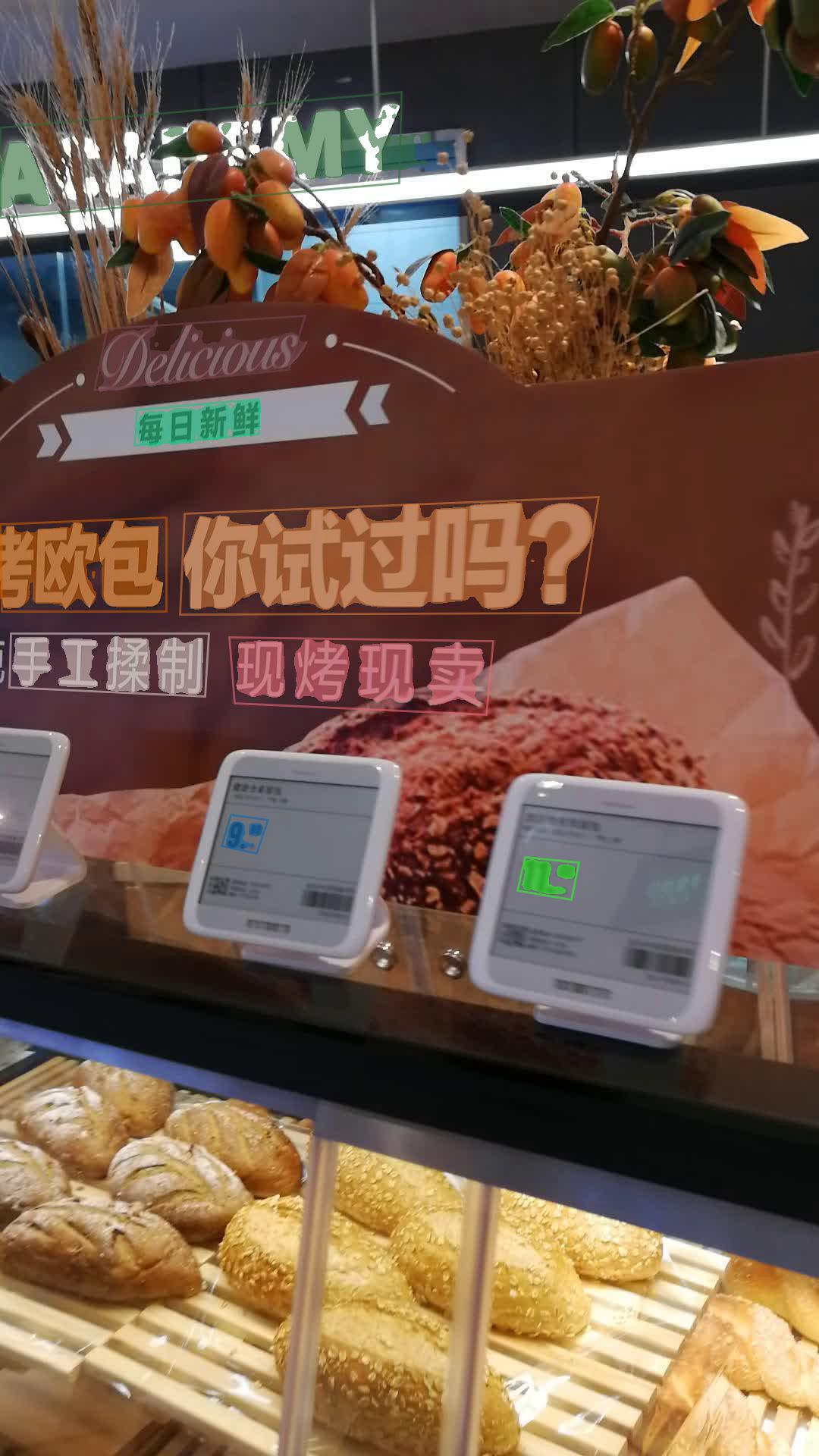}
  &\includegraphics[width=2.6cm,height=2.6cm]{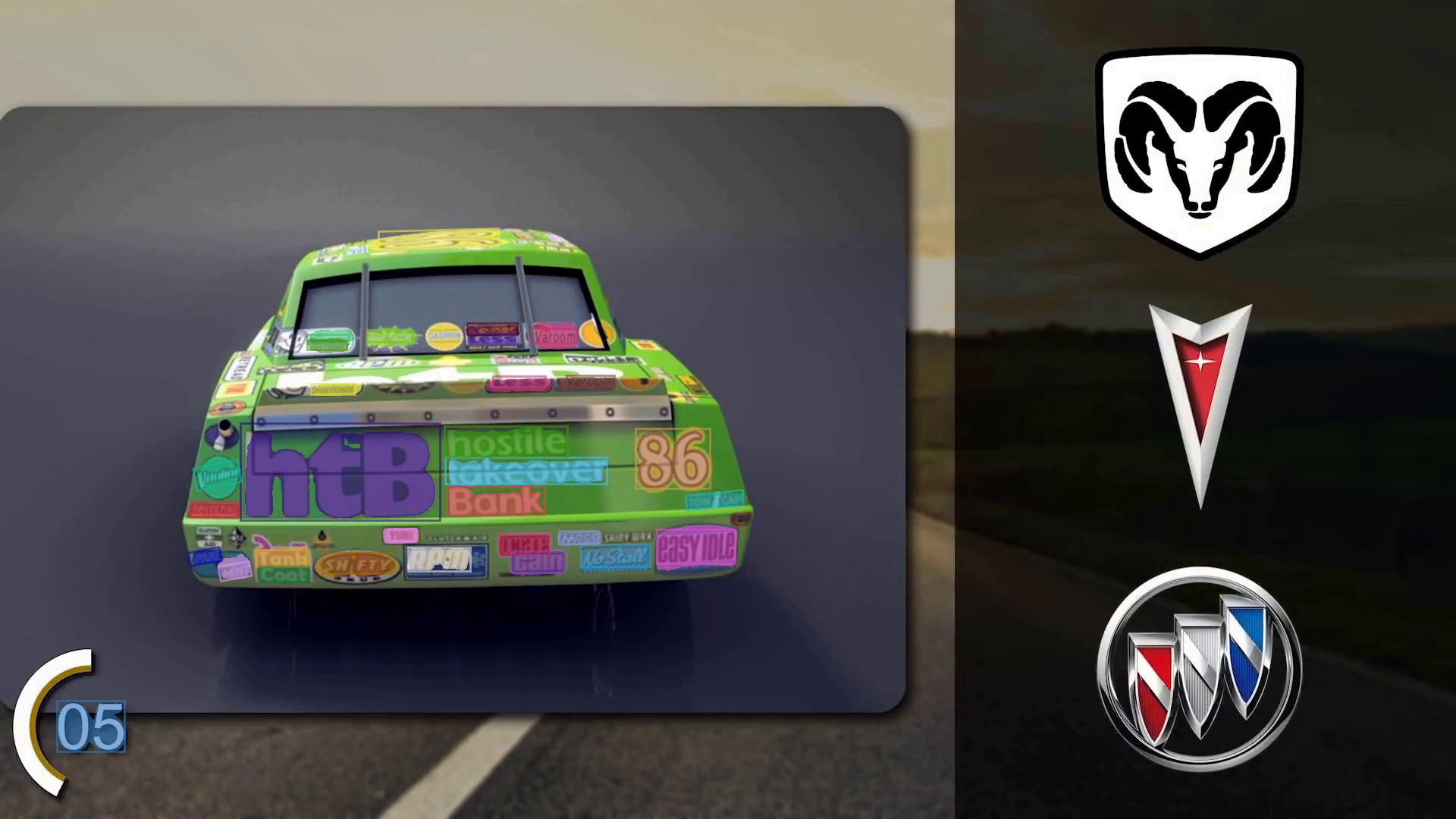}
  &\includegraphics[width=2.6cm,height=2.6cm]{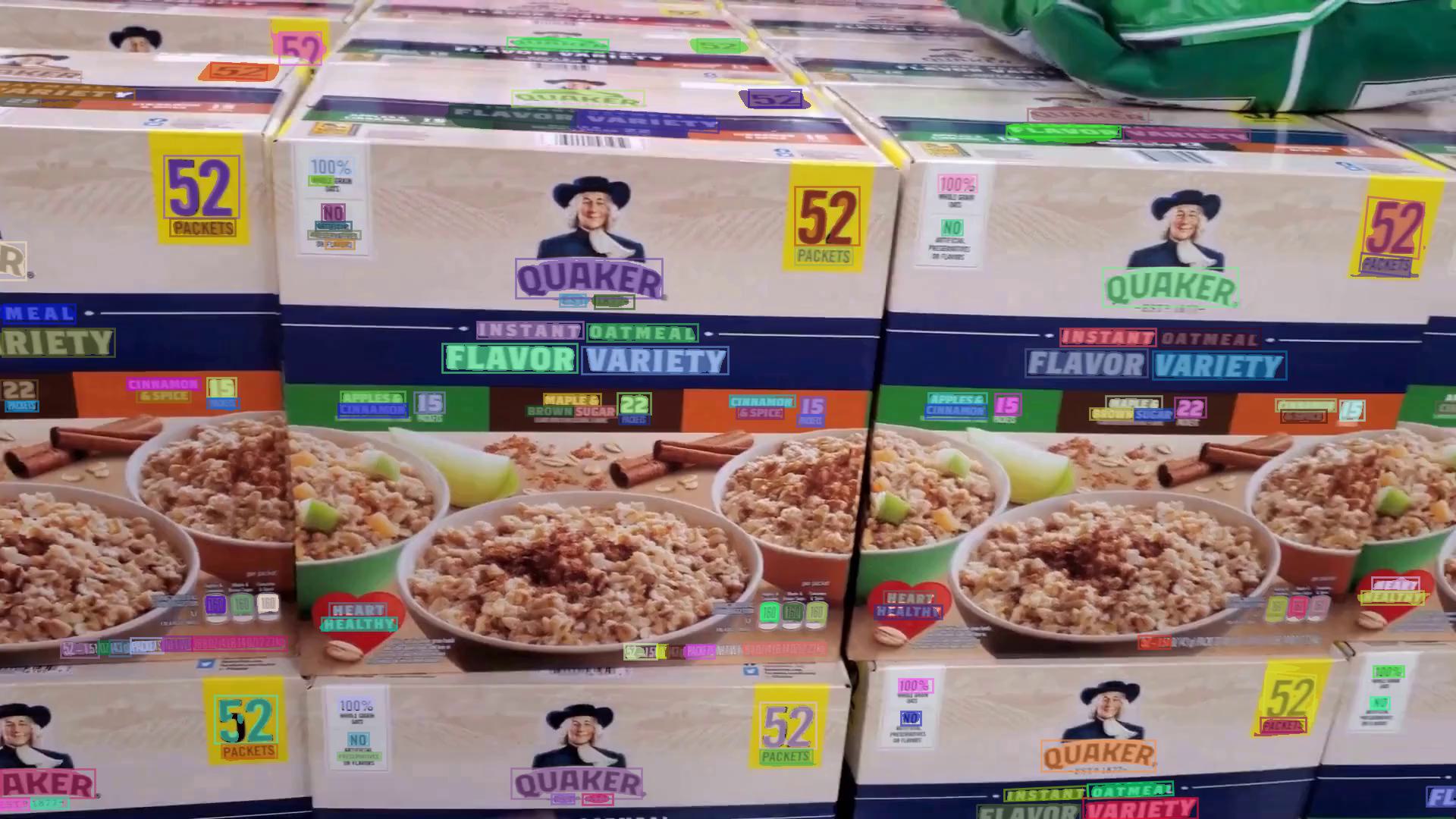}
  \\
  ICDAR15 &RoadText-1K &LSVDT &BOVText &DSText
  \\
  \end{tabular}
  \caption{Some visualization results of the generated masks in five datasets using the SAMText pipeline. The top row shows the scene text frames while the bottom row shows the generated masks.}
 \label{fig:2}
\end{figure}

\section{Results}

\begin{wrapfigure}[]{r}[0em]{0.5\textwidth}
\vspace{-10mm}
\begin{minipage}{1\linewidth}
\centering
\includegraphics[width=1\linewidth]{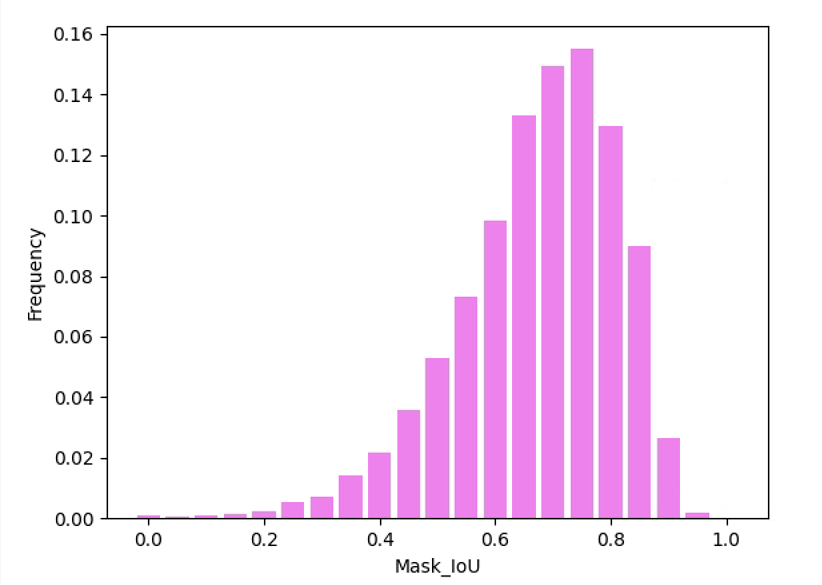} 
\end{minipage}
\caption{The distribution of IoU between the generated masks and ground truth masks in the COCO-Text training dataset~\cite{veit2016coco}.}
 \label{fig:mask_box_iou_all}
 \vspace{-10mm}
\end{wrapfigure}

\subsection{The Quality of Generated Masks} 
To evaluate the performance of SAMText, we select the COCO-Text training dataset \cite{veit2016coco} as it provides ground truth mask annotations for text instances. Specifically, we randomly sample 10\% of the training data and calculate the IoU between the masks generated by SAMText and their corresponding ground truth masks. Our findings show that SAMText has high accuracy, with an average IoU of 0.70. The histogram of IoU scores is shown in Fig.~\ref{fig:mask_box_iou_all}. Figure~\ref{fig:mask_box_iou_all} presents the histogram of IoU scores. Notably, the majority of IoU scores are centered around 0.75, suggesting that SAMText performs well.

\subsection{Visualization of Generated Masks}
In Figure~\ref{fig:2}, we show some visualization results of the generated masks in five datasets using the SAMText pipeline. The top row shows the scene text frames while the bottom row shows the generated masks. As can be seen, the generated masks possess fewer background components and align more precisely with the text boundaries than the bounding boxes. As a result, the generated mask annotations facilitate conducting more comprehensive research on this dataset, e.g., video text segmentation and video text spotting using mask annotations.

\begin{figure}[t]
\centering
\setlength{\tabcolsep}{0.1cm}
    \begin{tabular}{cc}
  \includegraphics[width=6.8cm,height=5.5cm]{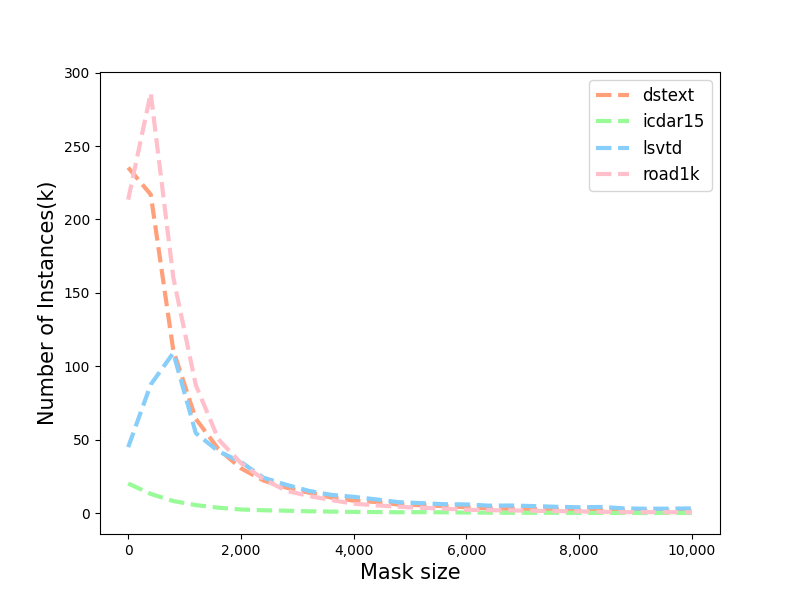} 
  &\includegraphics[width=6.8cm,height=5.5cm]{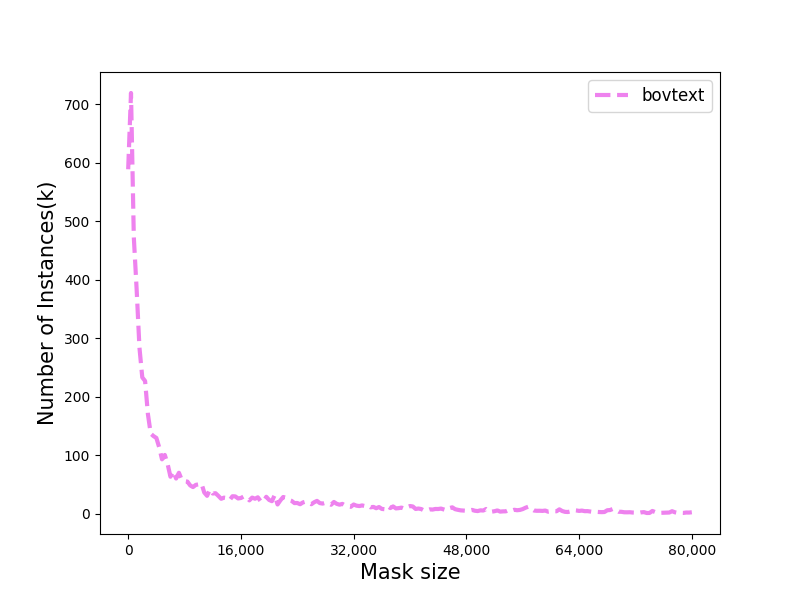}
  \\
  (a)  & (b)
    \end{tabular}
  \caption{(a) The mask size distributions of the ICDAR15, RoadText-1k, LSVDT, and DSText datasets. Masks exceeding 10,000 pixels are excluded from the statistics. (b) The mask size distributions of the BOVText datasets. Masks exceeding 80,000 pixels are excluded from the statistics.}
 \label{fig:3}
\end{figure}

\begin{figure}[t]
\centering
\setlength{\tabcolsep}{0.1cm}
    \begin{tabular}{cc}
  \includegraphics[width=6.8cm,height=5.5cm]{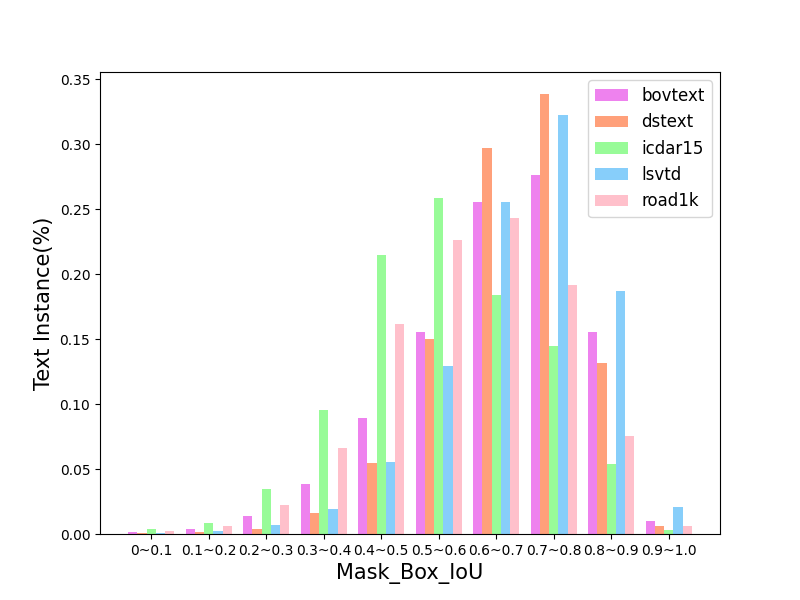} 
  &\includegraphics[width=6.8cm,height=5.5cm]{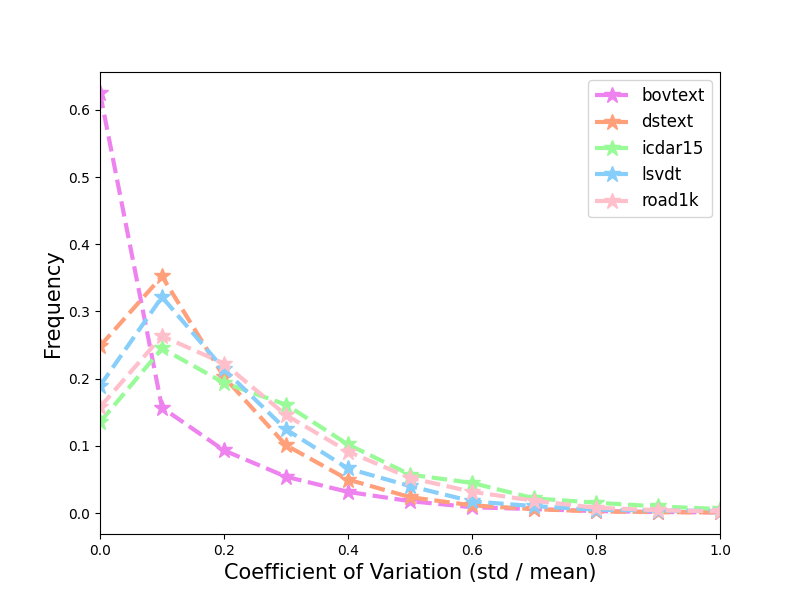}
  \\
  (a)  & (b)
    \end{tabular}
  \caption{(a) The distribution of IoU between the generated masks and ground truth bounding boxes in each dataset. (b) The CoV distribution of mask size changes for the same individual in consecutive frames in all five datasets, excluding the CoV scores exceeding 1.0 from the statistics.}
 \label{fig:4}
\end{figure}

\begin{figure}[t]
\centering
\setlength{\tabcolsep}{0.1cm}
\begin{tabular}{ccccc}
  \includegraphics[width=2.6cm,height=2.6cm]{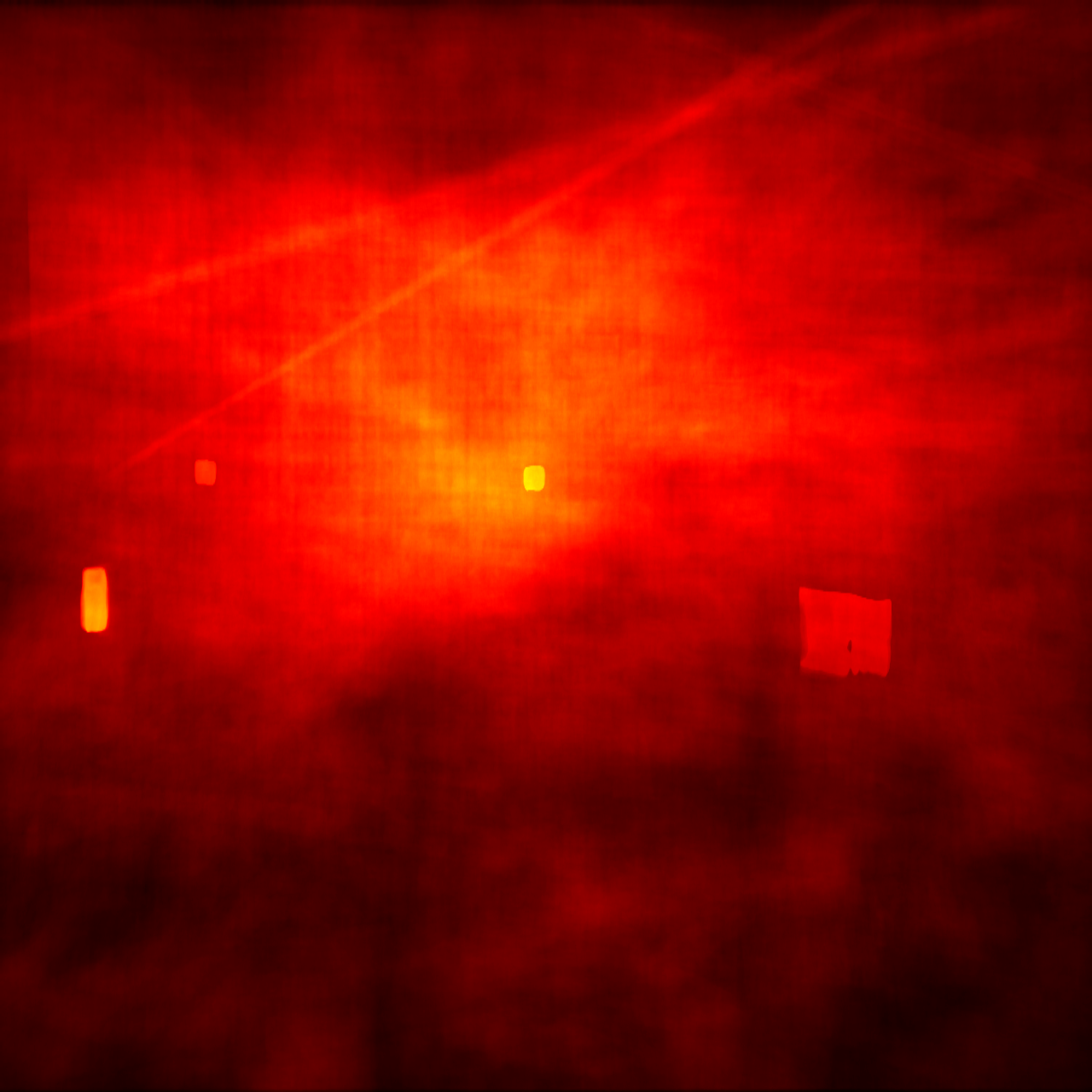} 
  &\includegraphics[width=2.6cm,height=2.6cm]{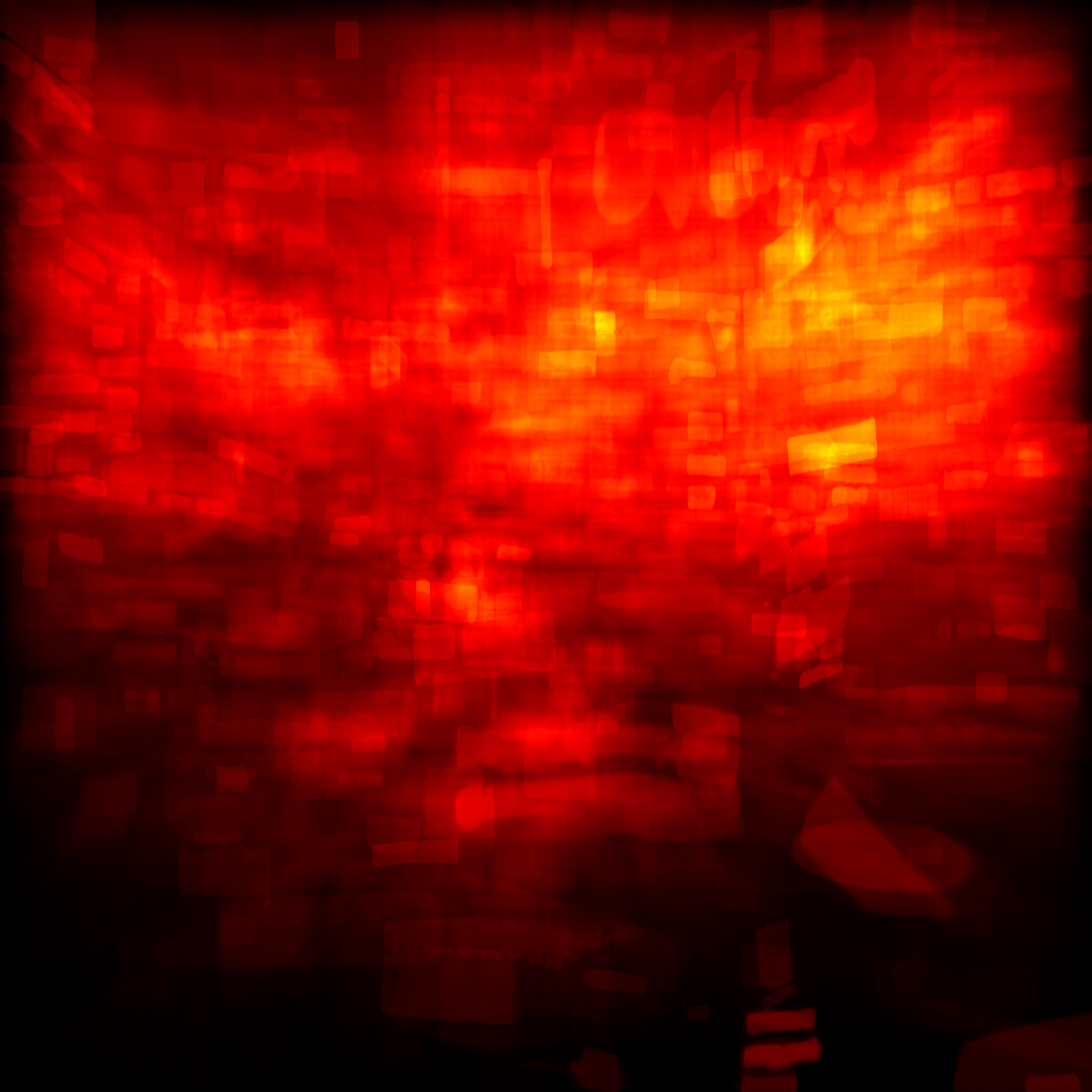}
  &\includegraphics[width=2.6cm,height=2.6cm]{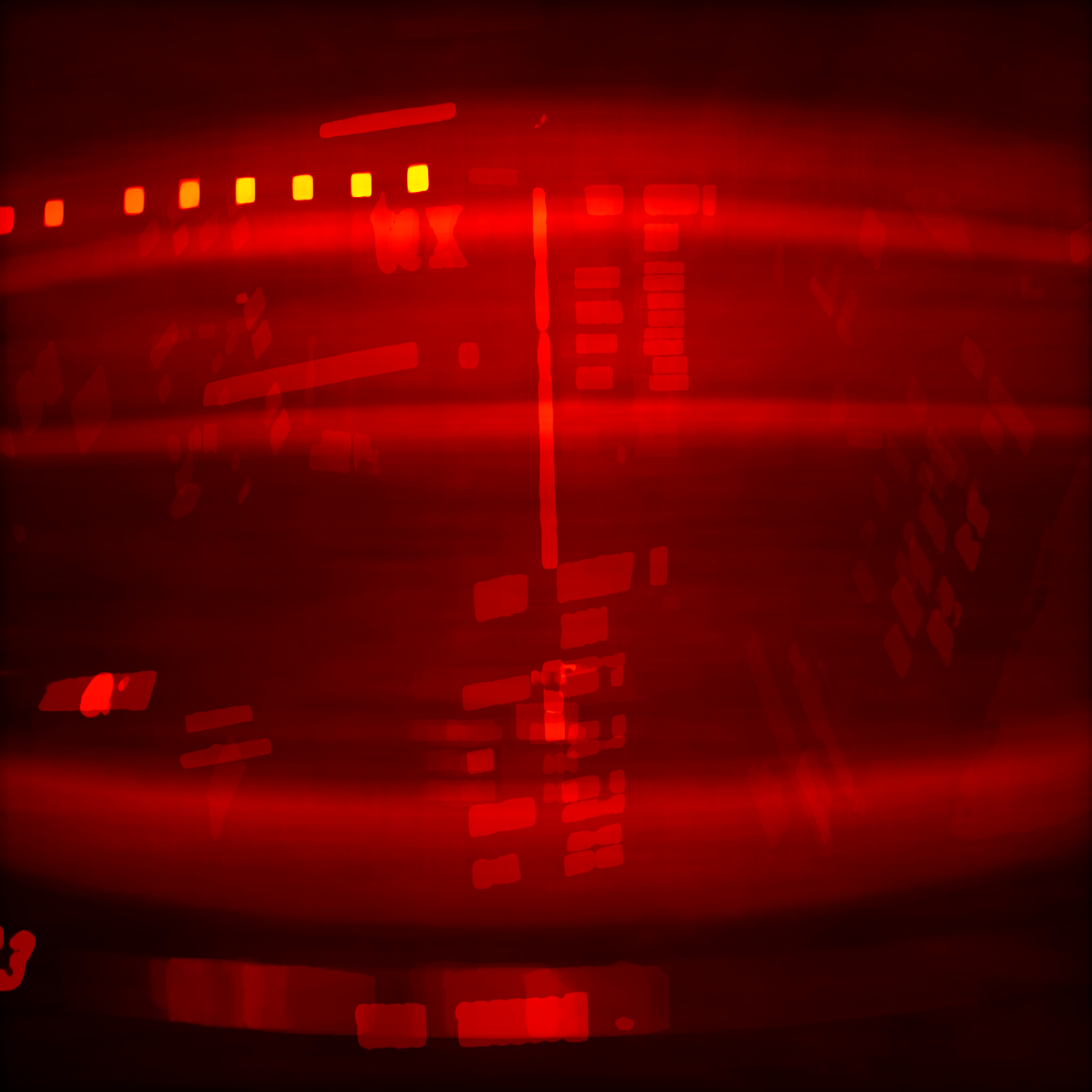}
  &\includegraphics[width=2.6cm,height=2.6cm]{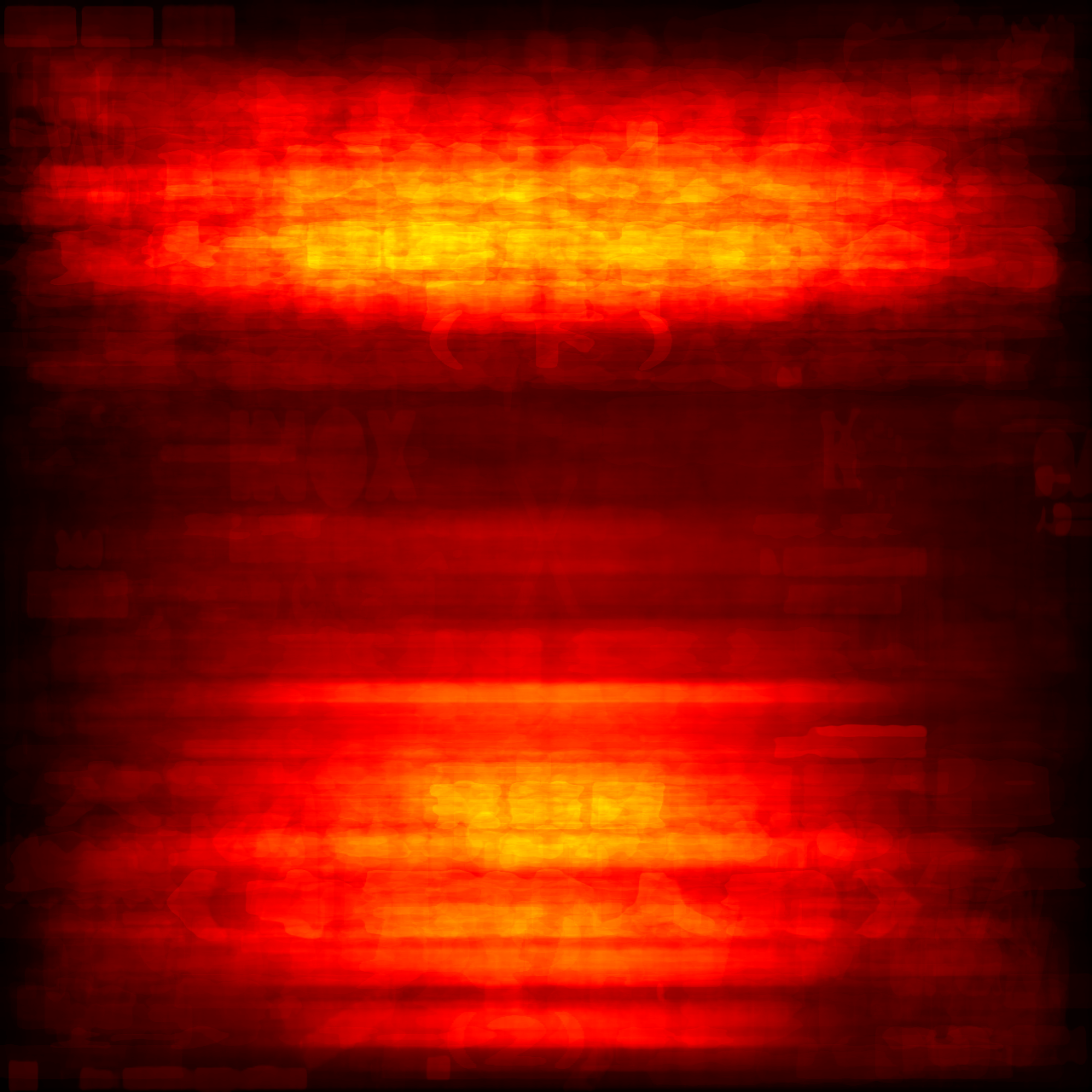}
  &\includegraphics[width=2.6cm,height=2.6cm]{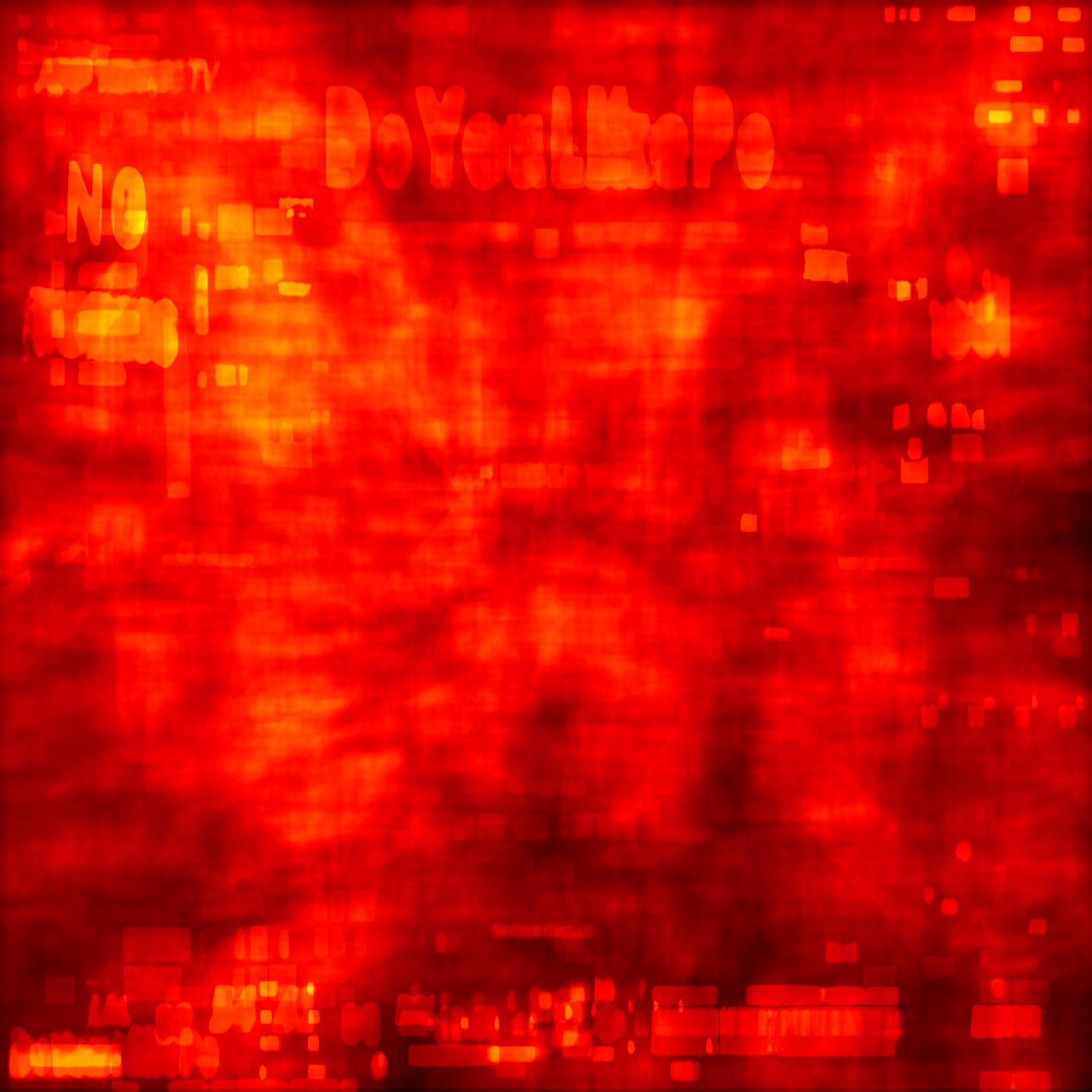} 
  \\ 
  ICDAR15 &RoadText-1K &LSVDT &BOVText &DSText
  \\
  \end{tabular}
  \caption{Visualization of the heatmaps that depict the spatial distribution of the generated masks in the five video text spotting datasets employed to establish SAMText-9M. Warmer colors indicate a higher frequency of mask occurrence.}
 \label{fig:5}
\end{figure}

\subsection{Dataset Statistics and Analysis}

\textbf{The size distribution.} As shown in Figure~\ref{fig:3}(a), the majority of mask sizes in the ICDAR15, RoadText-1k, LSVDT, and DSText datasets are less than 2,000 pixels, and their size distributions exhibit long tails. LSVDT contains fewer small instances (i.e., $<$2,000 pixels) than RoadText-1k and DSText, but more large ones. ICDAR15 exhibits smaller scales for both small and large instances than the other datasets. Conversely, as shown in Figure~\ref{fig:3}(b), BOVText comprises a more extensive range of instances than the aforementioned four datasets, with much larger mask sizes.

\textbf{The IoU distribution.} Figure \ref{fig:4}(a) displays histograms of the IoU scores between the masks generated by SAMText and the corresponding ground truth bounding boxes. The majority of IoU scores in BOVText, DSText, and LSVDT datasets exhibit peaks in the range of [0.7-0.8], whereas those of RoadText-1k and ICDAR15 datasets peak in the range of [0.6-0.7] and [0.5-0.6], respectively. The differences in the IoU distributions can be attributed to the blurry images in the RoadText-1k and ICDAR15, which are likely to have a side impact on the generated masks. It should be noted that the IoU scores between the generated masks by SAMText and their corresponding ground truth bounding boxes do not necessarily indicate the quality of the generated masks. Instead, they represent the ratio of foreground detected by SAMText to the background enclosed by the ground truth bounding boxes.

\textbf{The CoV distribution.} Figure~\ref{fig:4}(b) depicts the distribution of the CoV of the size of tracked instances across consecutive frames. The majority of instances show only small variations in different timestamps within the video clips, i.e., less than 10\%. However, there are many instances that demonstrate considerable size fluctuations, e.g., with CoV exceeding 40\%. This observation highlights a significant challenge for both text tracking and recognition tasks.

\textbf{The spatial distribution.} Figure~\ref{fig:5} shows the spatial distribution of text masks in various datasets. It is observed that the distribution of text masks in DSText is relatively homogeneous, while the masks in the remaining four datasets exhibit distinct spatial distributions. In particular, ICDAR15 and RoadText-1k share a similar pattern where masks are mainly situated in the upper region of the images. In contrast, masks in BOVText are frequently present in both the top and bottom portions of the images. The inclusion of all such instances in the SAMText-9M dataset can contribute significantly to the training of a more robust model.

\subsection{Promising Research Topics}
SAMText and SAMText-9M provide opportunities for exploring promising research topics in video text spotting. Some of these research topics are listed below:

\textbf{Impact of mask annotations.} With the availability of the extensive text masks in SAMText-9M, it is possible to design novel approaches for video text spotting that segment and track text instances. Meanwhile, it is worth exploring the benefits of using mask annotations over those based on HBB or OBB annotations. Besides, given that mask representation is more efficient than HBB and OBB in dealing with curved texts, another promising research direction is to investigate the impact of mask annotations for pre-training models designed for curved text spotting.

\textbf{Data scalability.} Improving the performance of video text spotting models with an increase in the amount of training data, i.e., achieving data scalability, is a crucial research topic in both academia and industry. The availability of a large number of text annotations in SAMText-9M makes it possible to investigate the ability of video text spotting models to learn complex patterns and generalize well to new data as the amount of training data increases, particularly for models that use vision transformer backbones~\cite{dosovitskiyimage,liu2021swin,xu2021vitae,zhang2023vitaev2,Related13}, which are known for their powerful representation capacity but require a large amount of training data to perform well. The SAMText pipeline only requires bounding boxes of text instances as input, which can be derived from a well-performed scene text detection model. Therefore, it is possible to generate mask annotations for text videos in the wild at scale, allowing for investigation of the impact of different amounts of unlabeled text videos for model training.

\textbf{Model scalability.} Similarly, model scalability is also a crucial aspect of deep learning, referring to the ability of models to scale up their parameters to improve their representation capacity and accommodate large-scale training data for enhanced performance. With the abundant annotations in SAMText-9M, it is worthwhile to explore the impact of large models, particularly those leveraging the rich supervisory signals from mask annotations, on the performance of video text spotting.

\textbf{Character-level annotation.} While SAMText can produce mask annotations for text instances, it struggles in distinguishing individual characters in each instance, particularly for blurry text or dense text with closely placed characters. Nevertheless, there is potential to improve SAMText to generate character-level mask annotations, which can be accomplished through various approaches, such as 1) sampling different points inside the text instance to prompt the SAM model to generate fine-grained masks, 2) fine-tuning a SAM model on training data with character-level mask annotations, particularly by utilizing the excellent few-shot learning capability of large models such as the SAM huge model, and 3) leveraging weak annotations, such as the number of characters in an instance, to train the segmentation model and improve its accuracy for character-level mask annotation.

\section{Conclusion}
This report introduces the SAMText pipeline, which offers an efficient and effective solution for scalable mask annotation in the field of video text spotting. Building upon this pipeline, we have created SAMText-9M, a new large-scale video text spotting dataset that includes about 9 million fine mask annotations. We have provided a comprehensive analysis of SAMText-9M's various statistics and analyzed its distinctive characteristics, such as its large scale and challenging nature. Furthermore, we have identified several promising research topics that warrant further exploration based on this dataset. We anticipate that SAMText and SAMText-9M will facilitate the advancement of video text spotting, and we intend to update this report as the project progresses.

\section*{Acknowledgment}
We acknowledge the authors of SAM for releasing the code and models, and the authors of the ICDAR15, RoadText-1K, LSVTD, BOVText, and DSText for releasing the datasets.

{
\small
\bibliographystyle{ieee}
\bibliography{ref}
}


\end{document}